# Explainable Label-flipping Attacks on Human Emotion Assessment System

Zhibo Zhang, Ahmed Y. Al Hammadi, Sangyoung Yoon, Ernesto Damiani, and Chan Yeob Yeun
C2PS, Department of Electrical Engineering and Computer Science, Khalifa University
Abu Dhabi, United Arab Emirates

*Abstract*—This paper's main goal is to provide an attacker's point of view on data poisoning assaults that use label-flipping during the training phase of systems that use electroencephalogram (EEG) signals to evaluate human emotion. To attack different machine learning classifiers such as Adaptive Boosting (AdaBoost) and Random Forest dedicated to the classification of 4 different human emotions using EEG signals, this paper proposes two scenarios of label-flipping methods. The results of the studies show that the proposed data poison attacks based on label-flipping are successful regardless of the model, but different models show different degrees of resistance to the assaults. In addition, numerous Explainable Artificial Intelligence (XAI) techniques are used to explain the data poison attacks on EEG signal-based human emotion evaluation systems.

## I. INTRODUCTION

The sources of industrial insider risk, which is a threat to an industrial organization, are always those who work there, such as current or past employees, and have inside knowledge of the business's security protocols, customer information, and computer systems. To assess human emotions and acts, speech [1] and facial expression [2] data were employed traditionally. With the correct training, people can, however, conceal voice and facial information with ease. On the other hand, EEG signals [3] have been utilized in recent years to assess a person's emotional state to prevent potential industrial insider assaults because people cannot conceal or manipulate their brainwaves.

Attackers could, however, find and take advantage of the flaws in machine learning models to lessen the efficiency of EEG signal-based human emotion assessment systems. Attackers contaminate a target Machine Learning model by poisoning it during the training phase using data poisoning (DP) [4] attack tactics. To particularly target the Machine Learning classifiers of EEG signal-based human emotion assessment systems, the purpose of this paper is to construct explainable label-flipping-based DP assaults. To particularly target the Machine Learning classifiers of EEG signal-based human emotion assessment systems, the purpose of this paper is to construct explainable label-flipping-based DP assaults [5]. The precise impact of DP attacks on EEG signal-based human emotion evaluation systems in terms of features and internal mechanisms is investigated and clarified using Explainable Artificial Intelligence (XAI) [6] techniques. Various poisoning thresholds have also been proposed in this research to quantify the poisoning impacts and vulnerabilities of each ML model.

Therefore, the primary contributions of this research article are as follows to cover the gap of deploying explainable label-flipping DP attacks against Machine Learning models of the EEG signal-based human emotion assessment systems from the attackers' perspective:

1) This paper applied different ML models in EEG signal-based risk evaluation systems and evaluated human emotions.

2) This research deployed a label-flipping attack in the training process of ML models.

3) This paper identified each model's vulnerabilities with the help of XAI techniques.

## II. METHOD AND RESULTS

The EEG signal dataset was gathered at Khalifa University in a specific facility in conformity with the ethical standards established by the Khalifa University Compliance Committee [7]. The dataset was compiled to look at potential applications of brainwave signals for spotting insider threats in the workplace. The Emotiv Insight 5 channels were the tool used to collect the data. Information from 17 people who gave their agreement to participate in the data collection is included in this dataset.

The four risk categories—High-Risk, Medium-Risk, Low-Risk, and Normal—found in the risk matrix were used to classify each signal for a captured image, and each signal was then given the appropriate label. The data files' 26 features are made up of 25 inputs and 1 output. Each of the five electrodes on the EEG gadget records one of the five brainwave bands: Theta, Alpha, Low Beta, High Beta, and Gamma.

Two different ML models, Random Forest and AdaBoost are intruded on by DP attacks. These ML classifiers were applied for the classification tasks of risk assessment based on the obtained EEG signal data without DP attacks initially. The dataset has been split into 80% for training the ML models, and 20% for testing. The proportion of poisoned samples to all samples is a crucial aspect to take into account when assessing the training set. Additionally, it is advantageous to evaluate the varied resilience capabilities of various ML models employed in the classification based on the EEG signal data from the attackers' perspective. Therefore, several poisoning rates were applied to the training set, such as 5%, 25%, 50%, and 75%, and the outcomes were compared to the models' starting performance (0% poisoning). Table 1 highlights the metrics comparison for the proposed DP attack under different poisoning rates and Fig. 1 shows the feature contributions under different poisoning rates using the XAI technique, Permutation Importance.

Fig. 1: Permutation Importance under label-flipping attack.

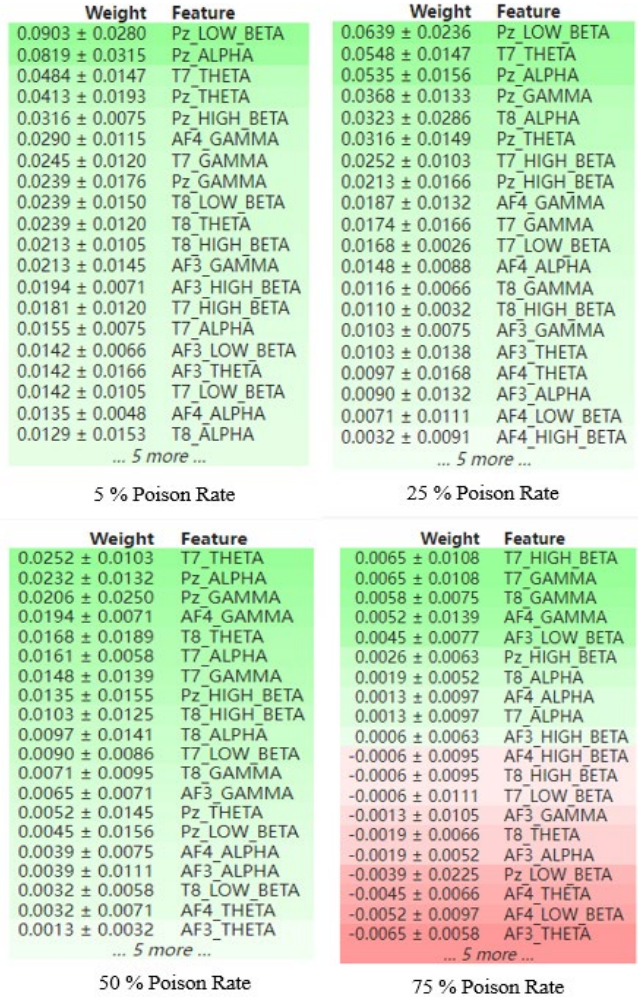

TABLE 1: Metrics Comparison among ML Models under DP Attack

| ML model | Poison rate [%] | Accuracy [%] | Recall [%] | Precision [%] | F1-Score [%] | Log loss |
|---|---|---|---|---|---|---|
| Ada Boost | 0 | 99.68 | 99.67 | 99.67 | 99.66 | 0.017 |
| | 5 | 96.77 | 96.78 | 96.79 | 96.74 | 0.118 |
| | 25 | 76.45 | 76.46 | 76.67 | 76.33 | 0.722 |
| | 50 | 57.74 | 58.25 | 63.59 | 56.61 | 1.799 |
| | 75 | 21.61 | 21.76 | 21.53 | 21.53 | 4.136 |
| Random Forest | 0 | 90.97 | 91.03 | 91.94 | 91.04 | 0.627 |
| | 5 | 88.71 | 88.85 | 89.87 | 88.79 | 0.659 |
| | 25 | 80.00 | 80.17 | 81.19 | 80.05 | 0.858 |
| | 50 | 50.00 | 50.56 | 57.41 | 48.02 | 0.995 |
| | 75 | 14.52 | 14.52 | 14.25 | 14.32 | 1.490 |

### III. Conclusion

The two categories of DP attacks based on Label Flipping on the ML models of the EEG Signal-based risk assessment systems were introduced and evaluated in this article. It has been demonstrated that both assaults can dramatically reduce the overall accuracy of classification rates for different multi-class risk assessment classifiers, although different ML classifiers exhibit diverse responses to various attacks. This study made use of several XAI tools to better assess the effects of the suggested two label-flipping attacks on various aspects of EEG brainwave signals and the outcomes of emotion classification. The results demonstrated that for classification and poisoning resilience, the AdaBoost classifier works best. Additionally, the Pz electrode's characteristics have a greater impact on the categorization outcomes of emotion assessment when using the XAI technique. Increasing the robustness of ML models against tampered training data to be used during re-training with the way of XAI methodology is another important target of this research's future study.